\documentclass[runningheads]{llncs}
\usepackage{graphicx}
\usepackage{amsmath}
\usepackage{amssymb}
\usepackage{amsfonts}
\usepackage{bm}
\usepackage{paralist}
\usepackage{float}
\usepackage{cite}
\usepackage{xcolor}
\usepackage{tabularx}
\usepackage{subcaption}
\usepackage{hyperref}
\usepackage{placeins}

\begin{document}
\title{Interpretable Machine Learning for TabPFN}
%
%
\author{David Rundel\inst{1} \and
Julius Kobialka\inst{1} \and
Constantin von Crailsheim\inst{1} \and
Matthias Feurer\inst{1,2} \and
Thomas Nagler\inst{1,2} \and
David Rügamer\inst{1,2}}
\authorrunning{Rundel et al.}
%
\institute{Department of Statistics, LMU Munich \and
Munich Center for Machine Learning
\email{\{firstname.lastname\}@stat.uni-muenchen.de}}
\maketitle              
\begin{abstract}
The recently developed Prior-Data Fitted Networks (PFNs) have shown very promising results for applications in low-data regimes. The TabPFN model, a special case of PFNs for tabular data, is able to achieve state-of-the-art performance on a variety of classification tasks while producing posterior predictive distributions in mere seconds by in-context learning without the need for learning parameters or hyperparameter tuning. This makes TabPFN a very attractive option for a wide range of domain applications. However, a major drawback of the method is its lack of interpretability. Therefore, we propose several adaptations of popular interpretability methods that we specifically design for TabPFN. By taking advantage of the unique properties of the model, our adaptations allow for more efficient computations than existing implementations. In particular, we show how in-context learning facilitates the estimation of Shapley values by avoiding approximate retraining and enables the use of Leave-One-Covariate-Out (LOCO) even when working with large-scale Transformers. In addition, we demonstrate how data valuation methods can be used to address scalability challenges of TabPFN. Our proposed methods are implemented in a package \texttt{tabpfn\_iml} and made available at \url{https://github.com/david-rundel/tabpfn\_iml}.
\end{abstract}

\section{Introduction}
Interpretable machine learning (IML) has become increasingly vital in the field of machine learning (ML), especially when applied to real-world tasks requiring an understanding of the model's predictions. IML serves both as a debugging tool to enhance fairness and ethics, as well as a means to improve social acceptance by fostering trust through interpretability \cite{molnar2020interpretable}.

Recently, Hollmann et al. \cite{hollmann2022tabpfn} introduced TabPFN, a prior-data fitted network (PFN) for tabular data that achieves state-of-the-art performance on a diverse range of classification tasks in mere seconds. It approximates probabilistic inference by in-context learning with a Transformer. For small tabular tasks, TabPFN shows great potential due to its straightforward use without requiring hyperparameter tuning. This renders TabPFN an appealing choice across diverse application domains. TabPFN, like other deep neural networks (DNNs), is however a black box and its missing interpretability makes it less attractive for applications in some research fields \cite{hollmann2022tabpfn, nagler2023statistical}. IML implementations have so far focused on classical supervised ML algorithms and are mostly model-agnostic. While there already exist model-specific implementations for certain deep learning (DL) and tree-based models that take advantage of their architectural properties \cite{molnar2020interpretable}, it is not always clear how to apply IML methods to PFNs and TabPFN in particular in a meaningful and efficient manner.
\paragraph{Our Contributions:}
In this work, we close this gap by modifying existing IML methods to take advantage of the unique properties of TabPFN and thereby enabling efficient computation of IML methods. In particular, these adaptations allow for a more accurate calculation of Shapley values through exact retraining and the use of LOCO, which is usually not a viable option for Transformers \cite{vaswani2017attention}. Also, we employ data valuation measures for context optimization, thereby mitigating scalability challenges of TabPFNs. Our methodological contribution is accompanied by a comprehensive, easy-to-use IML toolbox.
\section{Supervised Machine Learning \& TabPFN}
\label{sec:supervised_ml_tabpfn}
Consider $n_\text{train}$ independent and identically distributed (\textit{iid}) training observations, gathered in a training data set \(\mathcal{D}_P = \{\bm{x}^{(i)}, y^{(i)}\}_{i=1}^{n_\text{train}}\) where $\bm{x}^{(i)} \in \mathcal{X} \subseteq \mathbb{R}^p$ are $p$-dimensional feature vectors and $y^{(i)} \in \mathcal{Y} \subseteq \mathbb{R}^g$ are categorical or continuous targets. In the classical supervised ML setting, the goal is to learn the functional relationship between features and targets with a model $\hat{f}: \mathcal{X} \rightarrow \mathcal{Y}$ from a hypothesis space $\mathcal{H}$ that is fully determined by its parameters $\bm{\theta}$. For most ML algorithms, given the dataset and a hyperparameter configuration $\bm{\lambda} \in \Lambda$, a learner or inducer $\mathcal{I}(\mathcal{D}_P, \bm{\lambda})$ yields such a model by minimizing the empirical risk $\mathcal{R}_{\text{emp}}\bigl(\hat{f}(\{\bm{x}^{(i)}\}_{i=1}^{n_{\text{train}}}), \{y^{(i)}\}_{i=1}^{n_\text{train}}\bigr)$. This is accomplished by optimizing the parameters $\bm{\theta}$ that were either randomly initialized or pre-trained. The resulting model can be used to perform inference $\hat{f}(\{\bm{x}^{*(i)}\}_{i=1}^{n_{\text{inf}}})$ on an unseen inference dataset $\{\bm{x}^{*(i)}\}_{i=1}^{n_{\text{inf}}}$ of size $n_{\text{inf}}$ \cite{lundberg2017unified, bischl2023hyperparameter}. In the following, we focus on the binary classification task, i.e.,  $\mathcal{Y}= \{0, 1\}$, and a probabilistic model $\hat{f}(\cdot) \in [0,1]$ predicting the probability of the positive class given features and parameters $\bm{\theta}$. To highlight that $\hat{f}$ is fitted on the set of all available features $P= \{1,\dots, p\}$, we use $\hat{f}_P$ with index $P$. When the model is fitted on a feature subset $S \subseteq P$, giving rise to a reduced dataset $\mathcal{D}_S= \{\bm{x}^{(i)}_S, y^{(i)}\}_{i=1}^{n_{\text{train}}}$, where $\bm{x}^{(i)}= \{\bm{x}^{(i)}_S, \bm{x}^{(i)}_{S^{C}}\}$ for all $i=1, \dots, n_{\text{train}}$, we denote the model as $\hat{f}_S$.

For PFNs and TabPFNs \cite{muller2022transformers, hollmann2022tabpfn}, the training and inference procedure is, however, different. As opposed to classical supervised ML, where a new model is trained from scratch or fine-tuned on the training subset of a new dataset, PFNs perform inference in a single forward pass. In this approach, a pre-trained neural architecture with fixed parameters is fed with a training set $\mathcal{D}_P$ and inference features $\{\bm{x}^{*(i)}\}_{i=1}^{n_{\text{inf}}}$ from a previously unseen task. The model leverages the training set, also known as \textit{context}, to conduct inference on the inference observations (so-called \emph{in-context learning}). We denote the PFN model function as $\hat{f}_{\text{PFN}}(\{\bm{x}^{*(i)}\}_{i=1}^{n_{\text{inf}}}, \mathcal{D}_{P})$, emphasizing that both the inference and training data are supplied during inference. The index `\text{PFN}' is used to highlight that the parameters of PFNs remain unchanged with the training dataset. To achieve this, the underlying model has been pre-trained on artificially generated tasks sampled from a prior distribution. Hence, the training of PFNs consists of two steps: The first step is the offline training phase, where the model learns to approximate the posterior predictive distribution (PPD) for an inference observation on many synthetic training datasets from diverse tasks (about 10 million datasets in the case of TabPFN). In detail, for every synthetic dataset \(\mathcal{D}_{\text{syn}}\), it is trained to replicate the PPD on a query tuple \( ( Y_{\text{syn}}^{(n_{\text{syn}}+1)}, \bm{X}_{\text{syn}}^{(n_{\text{syn}}+1)} )\), using \(\mathcal{D}_{\text{syn}} \cup ( Y_{\text{syn}}^{(n_{\text{syn}}+1)}, \bm{X}_{\text{syn}}^{(n_{\text{syn}}+1)} )\) as training input. The second step is the inference phase, where a new training set and inference features are fed into the pre-trained network with fixed parameters \cite{hollmann2022tabpfn, nagler2023statistical}.

TabPFN specifically uses a Transformer model and a prior over tabular classification tasks. The tokens in the Transformer represent the observations, with training observations attending to each other and inference observations only attending to training observations. In particular, there are 
\(\binom{n_{\text{train}}}{2} + n_{\text{train}} \cdot n_{\text{inf}} 
\)
token connections and the computational cost for a forward pass scales as $\mathcal{O}(n_{\text{train}}^2 + n_{\text{train}} \cdot n_{\text{inf}})$. Due to the quadratic complexity in the number of training observations, the recommended maximum size of the training set is 1024 observations \cite{hollmann2022tabpfn, nagler2023statistical}. TabPFN eliminates the need for parameter training or hyperparameter tuning during the inference stage, enabling almost instantaneous predictions \cite{nagler2023statistical}. Several studies have demonstrated its effectiveness, achieving state-of-the-art performance across various benchmark datasets, specifically for moderately large tabular classification tasks \cite{mcelfresh2024neural, hollmann2022tabpfn}.
\section{Interpretable Machine Learning for TabPFN}
Existing IML implementations have primarily focused on traditional supervised ML algorithms or work for specific DNN architectures not meaningful or applicable to PFNs. In the following, we describe how to adapt existing IML methods to exploit the unique characteristics of TabPFN. This will render the methods more efficient while also preserving their conventional usage paradigms. Our methodological contributions and implementations include local and global methods for assessing feature effects (FEs), feature importance (FI), and data valuation (DV). We have selected the methods such that they complement each other w.r.t.\ their limitations. Where sensible, both continuous and categorical features are supported. Table~\ref{tab:methods} summarizes the methods and modifications made in this paper and implemented in the package \texttt{tabpfn\_iml}.
\begin{table}
\begin{center}
    \begin{tabular}{p{3cm}p{1cm}p{3cm}p{4cm}}
    {Method} & {Type} & {Scope} & {Modifications} \\
    \hline
    \hline
    ICE (\ref{subsection:3icepdpale}) & FE & local & Improved efficiency  \\
    PD (\ref{subsection:3icepdpale}) & FE & global & Improved efficiency  \\
    ALE (\ref{subsection:3icepdpale}) & FE & global & Improved efficiency  \\
    Kernel SHAP (\ref{subsection:3kernelshap}) & FE & local \& global & Exact retraining  \\
    SA (\ref{subsection:3sa}) & FE & local \& global & -  \\
    \hline
    LOCO (\ref{subsection:3loco}) & FI & global & Rendered tractable  \\
    SAGE (\ref{subsection:3further}) & FI & global & Exact retraining \\
    \hline
    LOO (\ref{subsection:3dvforco}) & DV & global & Rendered tractable  \\
    Data Shapley (\ref{subsection:3dvforco}) & DV & global & Exact retraining   \\
    SA (\ref{subsection:3sa}) & DV & local \& global & - \\
    \hline
    DCA (\ref{subsection:3further}) & CVA & global & - \\
    CP (\ref{subsection:3further}) & UQ & local & -  \\
    MOC (\ref{subsection:3further}) & LE & local & -  \\
    \hline
\end{tabular}
\vskip 0.1in
\caption{Overview of methods adapted and modified for TabPFN. In addition to FE, FI, and DV, we also include methods for clinical value assessment (CVA), uncertainty quantification (UQ), and local explanations (LE).}
    \label{tab:methods}
    \end{center}
    \vskip -0.2in
\end{table}
\subsection{ICE, PD and ALE} \label{subsection:3icepdpale}
Individual conditional expectation (ICE) curves \cite{molnar2020interpretable, goldstein2015peeking}, partial dependence (PD) plots \cite{molnar2020interpretable, friedman2001greedy} and accumulated local effect (ALE) plots \cite{molnar2020interpretable, apley2020visualizing} show how the predictions of a model change along a feature space of interest by systematically varying the respective input values. This necessitates prediction on artificial observations, where the feature of interest is varied according to a specified grid generated by discretizing the feature space into $G$ grid points. PD approximates the expected target response, while ALE addresses the extrapolation issue of ICE and PD \cite{molnar2020interpretable}.

Each TabPFN forward pass involves $n_{\text{train}} ((n_{\text{train}}-1)/2+ n_{\text{inf}})$ token connections (cf.~Section \ref{sec:supervised_ml_tabpfn}) and thus the computational cost per inference sample scales as $\mathcal{O}(n_{\text{train}}^2 / n_{\text{inf}})$. Consequently, TabPFN yields runtime improvements when processing as many inference samples as possible in a single forward pass. For the PD plot, widely used implementations such as the scikit-learn \texttt{partial\_dependence} function, execute a prediction step for every point of the grid. Instead, we propose a more efficient way to compute the PD plot in just a single forward pass by constructing a large inference array that contains all artificial observations. Compared to the scikit-learn implementation which has computational complexity $\mathcal{O}((n_{\text{train}}^2 + n_{\text{train}} \cdot n_{\text{inf}}) \cdot G)$, our implementation then scales as $\mathcal{O}(n_{\text{train}}^2 + n_{\text{train}} \cdot n_{\text{inf}} \cdot G)$. Thus, our PD plot implementation implies substantially lower runtimes. The modifications and implementations of ICE and ALE follow the same logic.
\subsection{LOCO} \label{subsection:3loco}
Leave-One-Covariate-Out (LOCO) estimates the effect of features on the predictive performance of a learner by retraining it on a subset of features without the feature of interest \cite{lei2018distribution, molnar2020interpretable}. The starting point is a model $\hat{f}_{P}$ trained with all features. Then, for each feature of interest $j$, the feature is removed from the dataset ($\mathcal{D}_{P\backslash \{j\}}$), the model is retrained without it ($\hat{f}_{P \backslash \{j\}}$) and the resulting change in empirical risk is tracked. LOCO approximately indicates whether a feature contains unique information that no other feature in the dataset contains \cite{lei2018distribution, molnar2020interpretable}. 

For many DL models, particularly extensive Transformer architectures, calculating LOCO scores is not feasible due to the high computational cost of repeatedly retraining models on different feature subsets. However, we can easily compute LOCO for TabPFN. Due to in-context learning, TabPFN enables inference on an updated training set through a single forward pass. In particular, we propose to calculate LOCO for the $j$-th feature as
\begin{align*}
\begin{split}
    \mathrm{LOCO}_j=\phantom{ }&
    \mathcal{R}_{\text{emp}}\bigl(
    \hat{f}_{\text{PFN}}(\{\bm{x}^{*(i)}_{P \backslash \{j\}}\}_{i=1}^{n_{\text{inf}}}, \mathcal{D}_{P \backslash \{j\}}), \{y^{*(i)}\}_{i=1}^{n_\text{inf}} \bigr)
    \\ &
    -
    \mathcal{R}_{\text{emp}}\bigl(
    \hat{f}_{\text{PFN}}(\{\bm{x}^{*(i)}_{P}\}_{i=1}^{n_{\text{inf}}}, \mathcal{D}_{P}), \{y^{*(i)}\}_{i=1}^{n_\text{inf}}\bigr).
\end{split}
\end{align*}
\subsection{Kernel SHAP} \label{subsection:3kernelshap}
Shapley values \cite{P-295} have their origin in game theory and aim to fairly attribute payouts. In ML, they estimate the local FE for an observation $\bm{x}^{*(i)}$ in a model $\hat{f}_P$ \cite{molnar2020interpretable}. In detail, the Shapley value for the $j$-th feature considers the (weighted) average marginal contribution of the feature to the model prediction across all feature subsets. While Shapley values provide a theoretically sound approach to attribution considering feature interactions, their calculation can be costly since the learner needs to be retrained for each feature subset $S$. As retraining ML models is in general resource-intensive and given that there are $2^p$ distinct feature subsets, this becomes impractical for most learners and datasets.

To address these challenges, Kernel SHapley Additive exPlanations (Kernel SHAP) \cite{lundberg2017unified} seeks to provide tractable approximations to Shapley values for a more efficient estimation. It employs an additive feature attribution measure by fitting a weighted linear model as local surrogate model that regresses model predictions on a binary representation of feature subsets, where only the features that appear in the subset are used for prediction \cite{molnar2020interpretable, lundberg2017unified}. To make the computations tractable, it incorporates the following additional approximations:
\begin{itemize}
    \item[\footnotesize $\bullet$] $M$ feature subsets are randomly sampled, as opposed to exhaustive consideration of all $2^p$ subsets. $M$ serves as a hyperparameter that balances computational cost and approximation quality \cite{vstrumbelj2014explaining, lundberg2017unified}.
    \item[\footnotesize $\bullet$] To avoid retraining the learner on every sampled feature subset $S$, retraining is approximated by utilizing the model trained on all features $\hat{f}_{P}$. This approach involves marginalizing over the features $S^C$ that are absent in a subset while assuming that all features are independent and approximating an integral by Monte Carlo.\footnote{In this work, we chose not to consider the optional model linearity assumption as it is overly simplistic.} Specifically, the approximation can be expressed as
    \begin{align}
            \hat{f}_{S}(\cdot)
            & \approx \frac{1}{L} \sum_{l=1}^{L} \hat{f}_{P}\bigl( (\, \cdot \,, \bm{x}^{\kappa(l)}_{S^{C}}) \bigr) \label{for:kshap_mcestim}.
    \end{align}   
    The indices $\kappa(l) \in \{1, \dots, n_{\text{train}}\}$ for $l \in \{1, \dots, L\}$ are randomly sampled and subsequently referred to as \textit{imputation samples}, used to impute $S^C$ \cite{vstrumbelj2014explaining, lundberg2017unified}.
\end{itemize} 
We can exploit the unique characteristics of TabPFN to enhance Kernel SHAP by obtaining exact solutions, rather than approximations, for certain terms in the Shapley value estimation. First, consider a na\"ive implementation of Kernel SHAP coupled with TabPFN. Using Equation~\ref{for:kshap_mcestim}, the approximation for retraining TabPFN on a feature subset $S$ is given by
    \begin{align*}
            \hat{f}_{\text{PFN}}(\{\bm{x}^{*(i)}_{S}\}_{i=1}^{n_{\text{inf}}}, \mathcal{D}_S) 
            & \approx \frac{1}{L} \sum_{l=1}^{L} \hat{f}_{\text{PFN}}(\{(\bm{x}^{*(i)}_{S}, \bm{x}^{\kappa(l)}_{S^{C}})\}_{i=1}^{n_{\text{inf}}}, \mathcal{D}_P).
    \end{align*}
An \textit{approximate retraining} implementation could process all $n_{\text{inf}}$ inference samples and $M$ feature subsets in parallel using $L$ imputation samples, scaling as
    \begin{align} \label{formula:tokenconsapprox}
    \mathcal{O}(n_{\text{train}}^2 + n_{\text{train}} \cdot n_{\text{inf}} \cdot M \cdot L).
    \end{align}
However, since training and prediction are done jointly in TabPFN's forward pass, $\hat{f}_{\text{PFN}}(\{\bm{x}^{*(i)}_{S}\}_{i=1}^{n_{inf}}, \mathcal{D}_S)$ does not need to be approximated. Instead, we iterate over $M$ feature subsets, for each restricting the features of both the training and inference set accordingly, and evaluate the model for all $n_{\text{inf}}$ inference samples. We refer to this approach as \textit{exact retraining} in our implementation, scaling as
    \begin{align} \label{formula:tokenconsexact}
    \mathcal{O}((n_{\text{train}}^2+ n_{\text{train}} \cdot n_{\text{inf}}) \cdot M).
    \end{align}
Assuming $n_{train} \approx n_{inf}$ and $L \geq 2$, our method has thus lower computational cost.
%
In Equation~\ref{formula:tokenconsapprox}, we assume that all calculations can be executed in a single forward pass. As this is mostly not possible due to memory constraints on commonly used hardware, we also provide an \textit{approximate retraining} implementation, where we iterate over $M$ and $L$, predicting sequentially in a memory-friendly manner.
\subsection{Data Valuation for Context Optimization}  \label{subsection:3dvforco}
One of the primary limitations of TabPFN is its scalability in $n_{\text{train}}$. As a consequence, its context is restricted to approximately 1000 observations. This limits the widespread adoption of TabPFN, especially for larger datasets \cite{feuer2024tunetables}.

A straightforward way for utilizing TabPFN with more extensive datasets involves employing random sketching. This entails selecting a random subset of size $n_{\text{sub}} \leq 1024$ from the complete training set as context. In contrast, context optimization involves employing more advanced strategies to determine appropriate contexts when dealing with large datasets. Recently, Feuer et al. \cite{feuer2024tunetables} introduced TuneTables, a prompt-tuning strategy for TabPFN that compresses datasets into learned contexts. In our work, we pursue a different approach to context optimization by selecting a representative training subset based on IML methods. Various data valuation techniques, e.g., Data Shapley, which quantify the contribution of training observations to the predictive performance, have been suggested in the literature. Empirical evidence further suggests that incorporating training samples based on data values can enhance performance of ML models \cite{ghorbani2019data}. 

In contrast to Shapley values for FEs, Data Shapley considers the marginal contributions of training observations -- as opposed to features  -- to the model performance across all training subsets. As retraining is impractical for complex learners and datasets, several approximations such as Gradient Shapley \cite{ghorbani2019data} have been proposed. Gradient Shapley employs simplistic approximations to estimate the marginal contributions, such as utilizing stochastic gradient descent with a batch size of one. Several advancements exist, but most of these involve additional approximations and yet require training multiple models, limiting their applicability \cite{kwon2023data, wang2023data, rozemberczki2022shapley}. We can, however, implement Data Shapley as a modification of Kernel SHAP with exact retraining by making use of in-context learning. This eliminates the reliance on such simplistic approximations. Consider a training dataset of size $n_{\text{train}} > n_{\text{sub}}$ (typically $n_{\text{sub}}= 1024$) where our objective is to identify an optimal subset of size $n_{\text{sub}}$. To achieve this, we evaluate $M$ randomly sampled training subsets, ranging in size from 256 (for sufficient context) to $n_{\text{sub}}$, on a validation set of size $n_{\text{val}}$.  The resulting empirical risk values are then attributed to the training set observations by fitting a weighted linear model as a local surrogate. Subsequently, we select the $n_{\text{sub}}$ training observations with lowest coefficients in the linear model as the context, expecting that these will contribute the least to the increase in empirical risk. Kernel SHAP forgoes the need to examine all $2^{n_{\text{train}}}$ training subsets. In order to ensure estimation of the surrogate model, we, however, require $M \geq n_{\text{train}}$, leading to substantial computational expenses. 

Furthermore, TabPFNs architecture facilitates the incorporation of additional DV techniques, including Leave-one-out (LOO) testing \cite{cook1977detection}.  Similar to LOCO, LOO removes a single training observation from the training procedure and computes the change in predictive performance \cite{wang2023data, rozemberczki2022shapley}. 
\subsection{Sensitivity Analysis}  \label{subsection:3sa}
Sensitivity Analysis (SA) is a model-specific IML method that leverages gradients computed in a backward pass to extract relevance scores for features \cite{gevrey2003review}. Computing SA for FE is straightforward for TabPFN and also included in our implementation.  Leveraging the differentiability of TabPFN w.r.t.\ training observations, we follow a similar approach to adapt SA for data values. For the data value of the $j$-th training observation, we compute the gradient 
\begin{align*}
R_{j}= \left \|\nabla_{(\bm x^{(j)}, y^{(j)})}
\mathcal{R}_{\text{emp}} \bigl( \hat{f}_{\text{PFN}}(\{\bm{x}^{*(i)}\}_{i=1}^{n_{\text{inf}}}, \mathcal{D}_{P}), \{y^{*(i)}\}_{i=1}^{n_\text{inf}} \bigr)
\right\|_2
\end{align*}
of the empirical risk w.r.t.\ that observation and aggregate it to a scalar using the L2-norm. This allows identifying training observations for which the empirical risk is highly sensitive.
\subsection{Further Methods}  \label{subsection:3further}
In order to complement our toolbox to provide a comprehensive IML tool, we include Shapley Additive Global Importances (SAGE) \cite{NEURIPS2020_c7bf0b7c} as an additional FI method. Our implementation involves utilizing Kernel SHAP with the empirical risk as the function to be explained and also benefits from exact retraining. We also offer wrappers for decision curve analysis (DCA), a CVA method as implemented in \cite{sjoberg2022dcurves}, and conformal prediction (CP), an UQ method as implemented in \cite{taquet2022mapie}. Last but not least, we implement multi-objective counterfactual explanations (MOC) from the realm of LE using evolutionary multi-objective algorithms (EMOAs) \cite{dandl2020multi}. For more information, we refer to the \href{https://github.com/david-rundel/tabpfn_iml}{\texttt{tabpfn\_iml} repository}.
\section{Numerical Experiments\protect\footnote{All experiments were run on a Nvidia GTX A6000 GPU. Furthermore, our experiment code in the repository enables the replication of all results.}}
In the following, we provide numerical evidence of the improvement in computational efficiency and approximation error obtained with our modifications and implementation. Furthermore, we show how DV methods can be employed to improve classification performance by context optimization.
\begin{figure}[t]
\centering
\begin{subfigure}{.5\textwidth}
  \centering
  \includegraphics[width=\linewidth]{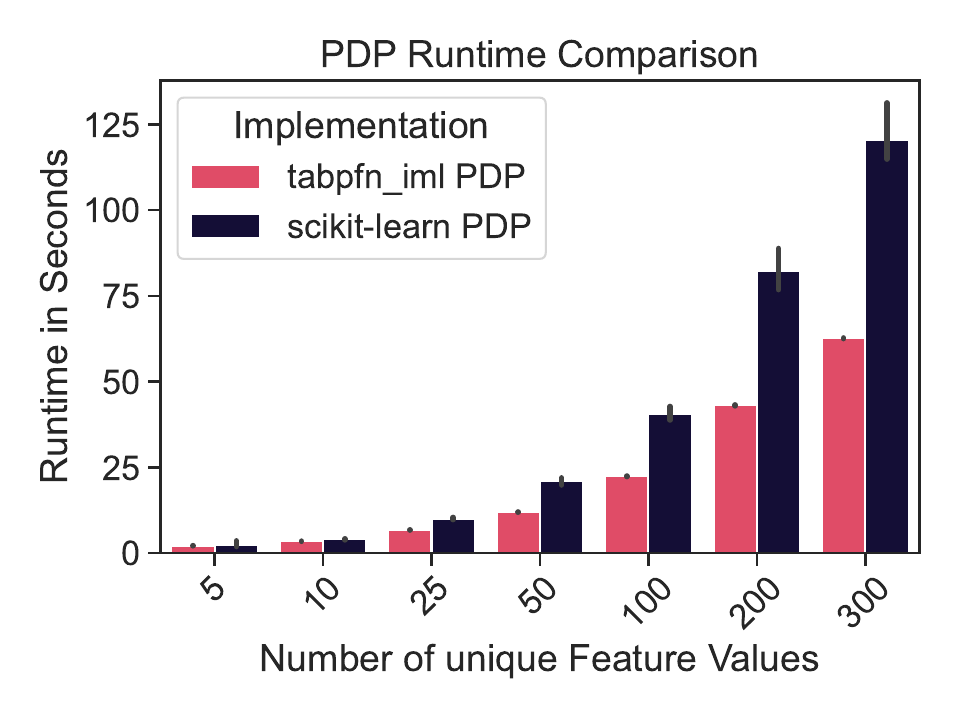}
  \caption{Synthetic datasets}
  \label{fig:pdp_synth}
\end{subfigure}%
\begin{subfigure}{.5\textwidth}
  \centering
  \includegraphics[width=\linewidth]{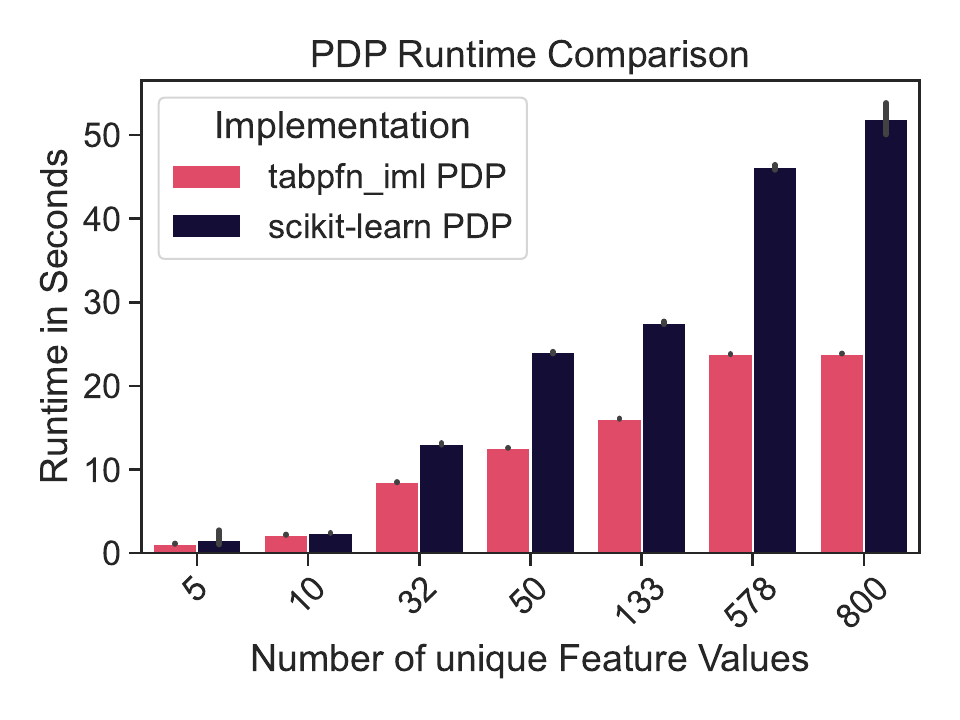}
  \caption{OpenML-CC18 repository datasets}
  \label{fig:pdp_openml}
\end{subfigure}
\caption{Runtime comparison of the implementations of \texttt{tabpfn\_iml} PD vs.~scikit-learn \texttt{partial\_dependence()}.}
\label{fig:pdp}
\end{figure}
%
\subsection{ICE \& PD}
To show the improved efficiency of our ICE and PD implementation, we conduct a runtime comparison across 25 synthetic datasets with 10 features and 1000 observations in Figure \ref{fig:pdp_synth} and across 7 binary classification datasets with increasing size from the OpenML-CC18 repository in Figure \ref{fig:pdp_openml}.
For the synthetic data, we discretize the feature of interest to the number of intervals indicated on the $x$-axis. We compare our implementation to scikit-learn's \texttt{partial\_dependence()} across five different seeds per feature level, setting the \texttt{grid\_size} equal to the number of unique feature values. In Figure \ref{fig:pdp_openml}, where runtimes for different real-world datasets are depicted, we set the \texttt{grid\_size} parameter to the default value of 100. For both Figures, $n_{\text{train}}$ comprises 80\% of all observations, and $n_{\text{inf}}$ the remaining 20\%. Our modifications to ICE and PD result in much better runtime, with our implementation being substantially faster as the grid size or data set size increase.
\subsection{Kernel SHAP}
\begin{figure}[t]
    \centering
    \includegraphics[width=\linewidth]{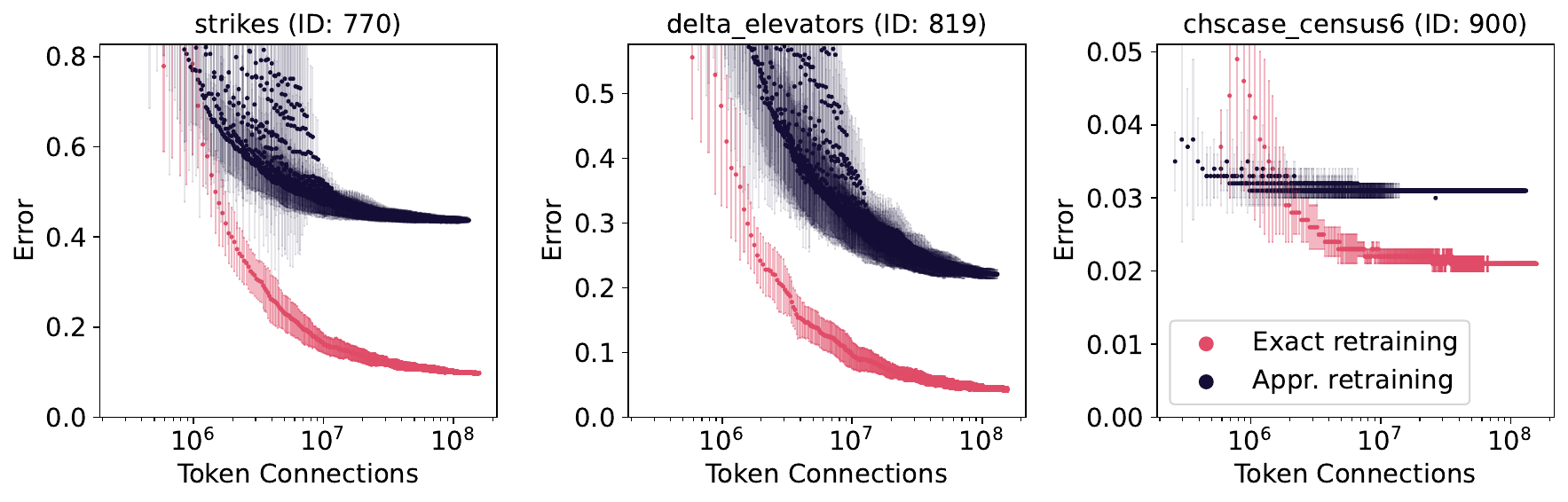}
    \caption{Average errors (points) and corresponding standard deviations (error bars) across all runs plotted against the token connections for various configurations and three datasets (different plots). Exact retraining configurations are red, while approximate retraining configurations are black.}
    \label{fig:error_tc}
\end{figure}
Next, we empirically compare our proposed exact retraining method to the alternative approximate method. The comparison of the two Kernel SHAP approaches is done by defining a (stochastic) error metric, which for each calculates the sum of the absolute deviations from the exactly computed Shapley values $\phi_{j}(\hat{f}_{\text{PFN}}, \bm{x}^{*(i)})$ across features $j$, averaged over inference observations $i$:
\begin{align*}
    \frac{1}{n_{\text{inf}}}
    \sum_{i=1}^{n_{\text{inf}}} \sum_{j=1}^{P}  \quad \lvert
    \phi_{j}(\hat{f}_{\text{PFN}}, \bm{x}^{*(i)})
    - 
    \hat{\phi}_{j, {\textit{(M,L)}}}(\hat{f}_{\text{PFN}}, \bm{x}^{*(i)})
    \rvert.
\end{align*}
$\hat{\phi}_{j, {\textit{(M,L)}}}(\hat{f}_{\text{PFN}}, \bm{x}^{*(i)})$ indicates the estimated Shapley value for the $j$-th feature and $i$-th inference observation. We further specify $M \in \{1, 2, \dots \}$ and $L \in \{-1, 1, 2, \dots \}$ as the number of feature subsets and imputation samples, where we denote exact retraining as $L = -1$.
Since the exact computation of Shapley values scales exponentially in the number of features, we select three binary classification tasks from the OpenML-CC18 repository \cite{OpenML2021} with six numerical features, aligning with the optimal application scenario for TabPFN \cite{hollmann2022tabpfn}. For each dataset, we randomly sample a training subset of size $n_{\text{train}}= 256$ and an inference set of size $n_{\text{inf}}= 128$ and evaluate a fine-grained grid across all values of $M \in \{6, \dots, 160\}$ and $L \in \{1, \dots, 25\}$ for approximate retraining, and across values of $M \in \{6, \dots, 1600\}$ for exact retraining. We repeat all experiments 25 times for each dataset, with feature subsets and imputation samples randomly sampled in each iteration.

In Figure \ref{fig:error_tc}, we present the mean and one standard deviation of the error across all runs, contrasted with the corresponding number of token connections for each $(M,L)$ configuration according to Equations~\ref{formula:tokenconsapprox} and \ref{formula:tokenconsexact}. The plot shows that exact retraining almost exclusively achieves lower approximation error for a fixed computational budget. In contrast, achieving comparable error values with approximate retraining often demands a larger number of feature subsets and multiple imputation samples. Moreover, exact retraining converges at a lower error and the standard deviations indicate that its Shapley estimates have lower variance. On top, it is important to stress that the number of token connections for approximate retraining is a lower bound and approximate retraining converges even slower in practice. Notably, for the chscase\_census6 dataset, there are configurations with a small number of token connections where approximate retraining outperforms the exact equivalent, which might be attributed to the cost-effectiveness of approximate retraining for low values of $L$.
\subsection{Data valuation for Context Optimization}
\begin{figure}[t]
    \centering
    \includegraphics[width=\linewidth]{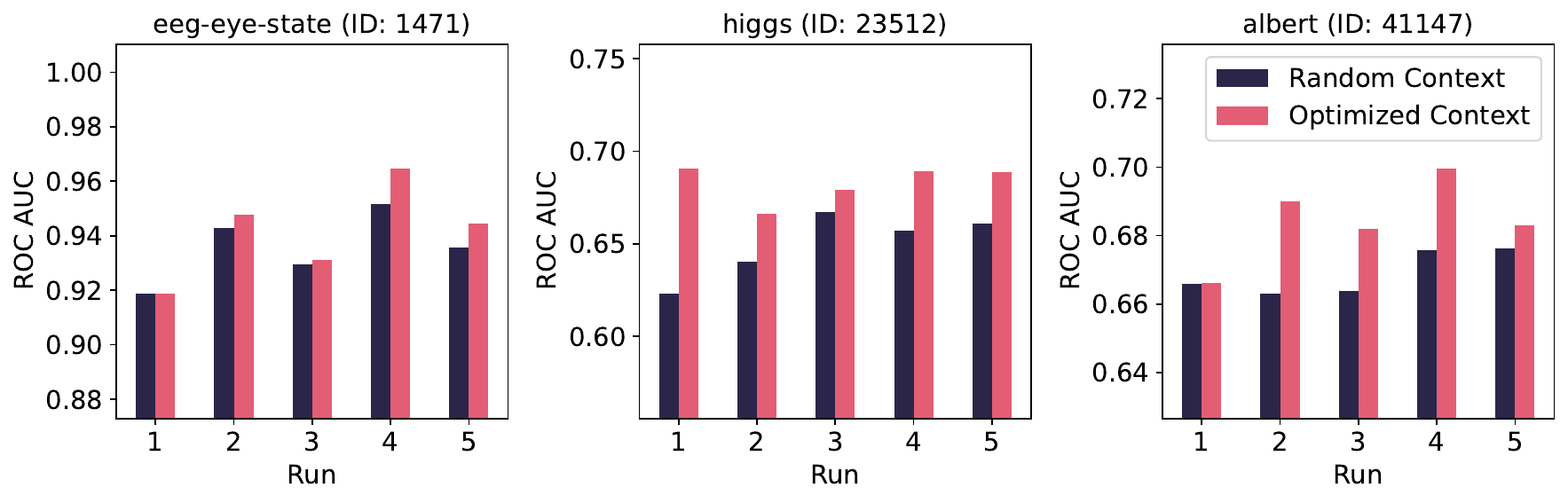}
    \caption{Comparison of the ROC AUC of randomly selected TabPFN contexts (black) and contexts optimized through Data Shapley coupled with Kernel SHAP (red) across five runs and three datasets.}
    \label{fig:rocauc_run}
\end{figure}
We conduct an empirical evaluation of our implementation of Data Shapley for context optimization by comparing it to random sketching. Our benchmark consists of three datasets from the OpenML-CC18 benchmarking suite \cite{OpenML2021}, chosen in accordance with \cite{feuer2024tunetables}, suggesting that TabPFN's performance on these datasets can be enhanced through context optimization. For each dataset, we create a training set of $n_{\text{train}}=3072$ observations and aim to identify an optimal context of size $n_{\text{sub}}=512$ thereof. We choose $M= 3 \cdot n_{\text{train}}$, $n_{\text{val}}= 512$ and ultimately assess the optimized context's performance on a previously unseen test set containing 1024 instances, comparing it to the performance of randomly selected contexts of the same size. This procedure is repeated 5 times with differently sampled training sets and subsets. 

In Figure \ref{fig:rocauc_run}, we compare the ROC AUC on the test set of both random and optimized contexts across all runs and three datasets. The plot indicates that our method consistently outperforms random sketching. On the eeg-eye-state and albert datasets, we observe an average increase in ROC AUC of 0.57\% and 1.52\%, respectively. An even more substantial improvement is obtained for the higgs dataset, yielding a difference of 3.3\% in ROC AUC. 

The results suggest that our implementation of Data Shapley produces sensible results, indicating its potentially meaningful application in domains beyond context optimization. However, note that this procedure requires nearly 10,000 TabPFN forward passes and is -- in contrast to our other proposed methods -- notably more expensive.
\section{Discussion \& Conclusion}
By exploiting the architecture underlying TabPFN and in-context learning, we propose tailored modifications and adaptations of IML methods for TabPFN. Specifically, we provide efficient implementations of ICE, PD, and ALE. Furthermore, using similar insights, we make the computation of LOCO and LOO possible despite TabPFN's Transformer nature. Moreover, we have shown that our exact retraining approach to Kernel SHAP dominates current alternatives with respect to both approximation error and runtime. Finally, our implementation of Data Shapley yields meaningful results that can be used for context optimization. Together with further methods in IML, we provide a comprehensive and easy-to-use IML package that allows to shed light on TabPFN's predictions. Exploring the potential of the suggested modifications to IML methods in the broader realm of in-context learning is a promising direction for future research.
\bibliographystyle{splncs04}
\bibliography{references}

\begin{thebibliography}{10}
\providecommand{\url}[1]{\texttt{#1}}
\providecommand{\urlprefix}{URL }
\providecommand{\doi}[1]{https://doi.org/#1}

\bibitem{apley2020visualizing}
Apley, D.W., Zhu, J.: Visualizing the effects of predictor variables in black box supervised learning models. J.R. Stat. Soc.  \textbf{82}(4),  1059--1086 (2020)

\bibitem{bischl2023hyperparameter}
Bischl, B., Binder, M., Lang, M., Pielok, T., Richter, J., Coors, S., Thomas, J., Ullmann, T., Becker, M., Boulesteix, A.L., et~al.: Hyperparameter optimization: Foundations, algorithms, best practices, and open challenges. Data Min. Knowl. Discov.  \textbf{13}(2),  e1484 (2023)

\bibitem{OpenML2021}
Bischl, B., Casalicchio, G., Feurer, M., Gijsbers, P., Hutter, F., Lang, M., Gomes~Mantovani, R., van Rijn, J., Vanschoren, J.: Openml benchmarking suites. In: NeurIPS Systems Track on Datasets and Benchmarks. vol.~1 (2021)

\bibitem{cook1977detection}
Cook, R.D.: Detection of influential observation in linear regression. Technometrics  \textbf{19}(1),  15--18 (1977)

\bibitem{NEURIPS2020_c7bf0b7c}
Covert, I., Lundberg, S.M., Lee, S.I.: Understanding global feature contributions with additive importance measures. In: NeurIPS. vol.~33, pp. 17212--17223 (2020)

\bibitem{dandl2020multi}
Dandl, S., Molnar, C., Binder, M., Bischl, B.: Multi-objective counterfactual explanations. In: PPSN XVI, Proceedings, Part I. pp. 448--469. Springer (2020)

\bibitem{feuer2024tunetables}
Feuer, B., Schirrmeister, R.T., Cherepanova, V., Hegde, C., Hutter, F., Goldblum, M., Cohen, N., White, C.: Tunetables: Context optimization for scalable prior-data fitted networks. arXiv:2402.11137  (2024)

\bibitem{friedman2001greedy}
Friedman, J.H.: Greedy function approximation: a gradient boosting machine. Annals of statistics pp. 1189--1232 (2001)

\bibitem{gevrey2003review}
Gevrey, M., Dimopoulos, I., Lek, S.: Review and comparison of methods to study the contribution of variables in artificial neural network models. Ecological modelling  \textbf{160}(3),  249--264 (2003)

\bibitem{ghorbani2019data}
Ghorbani, A., Zou, J.: Data shapley: Equitable valuation of data for machine learning. In: ICML. PMLR, vol.~97, pp. 2242--2251 (2019)

\bibitem{goldstein2015peeking}
Goldstein, A., Kapelner, A., Bleich, J., Pitkin, E.: Peeking inside the black box: Visualizing statistical learning with plots of individual conditional expectation. J. Comput. Graph. Stat.  \textbf{24}(1),  44--65 (2015)

\bibitem{hollmann2022tabpfn}
Hollmann, N., M{\"u}ller, S., Eggensperger, K., Hutter, F.: {TabPFN: A Transformer That Solves Small Tabular Classification Problems in a Second}. In: ICLR (2023)

\bibitem{kwon2023data}
Kwon, Y., Zou, J.: {Data-OOB: out-of-bag estimate as a simple and efficient data value}. In: ICML. PMLR, vol.~202, pp. 18135--18152 (2023)

\bibitem{lei2018distribution}
Lei, J., G’Sell, M., Rinaldo, A., Tibshirani, R.J., Wasserman, L.: Distribution-free predictive inference for regression. J. Am. Stat. Assoc.  \textbf{113}(523),  1094--1111 (2018)

\bibitem{lundberg2017unified}
Lundberg, S.M., Lee, S.I.: A unified approach to interpreting model predictions. In: NeurIPS. vol.~30 (2017)

\bibitem{mcelfresh2024neural}
McElfresh, D., Khandagale, S., Valverde, J., Prasad~C, V., Ramakrishnan, G., Goldblum, M., White, C.: When do neural nets outperform boosted trees on tabular data? In: NeurIPS. vol.~36 (2024)

\bibitem{molnar2020interpretable}
Molnar, C.: Interpretable machine learning. 2 edn. (2020)

\bibitem{muller2022transformers}
M{\"u}ller, S., Hollmann, N., Arango, S.P., Grabocka, J., Hutter, F.: Transformers can do bayesian inference. In: ICLR (2022)

\bibitem{nagler2023statistical}
Nagler, T.: Statistical foundations of prior-data fitted networks. In: ICML. PMLR, vol.~202, pp. 25660--25676 (2023)

\bibitem{rozemberczki2022shapley}
Rozemberczki, B., Watson, L., Bayer, P., Yang, H.T., Kiss, O., Nilsson, S., Sarkar, R.: {The Shapley Value in Machine Learning}. In: {IJCAI}. pp. 5572--5579 (2022)

\bibitem{P-295}
Shapley, L.S.: A Value for N-Person Games. RAND Corporation, Santa Monica, CA (1952)

\bibitem{sjoberg2022dcurves}
Sjoberg, D.D.: dcurves: Decision Curve Analysis for Model Evaluation (2022), https://github.com/ddsjoberg/dcurves

\bibitem{vstrumbelj2014explaining}
{\v{S}}trumbelj, E., Kononenko, I.: {Explaining Prediction Models and Individual Predictions with Feature Contributions}. Knowledge and information systems  \textbf{41},  647--665 (2014)

\bibitem{taquet2022mapie}
Taquet, V., Blot, V., Morzadec, T., Lacombe, L., Brunel, N.: {MAPIE: An Open-Source Library for Distribution-Free Uncertainty Quantification}. arXiv:2207.12274  (2022)

\bibitem{vaswani2017attention}
Vaswani, A., Shazeer, N., Parmar, N., Uszkoreit, J., Jones, L., Gomez, A.N., Kaiser, {\L}., Polosukhin, I.: Attention is all you need. In: NeurIPS. vol.~30 (2017)

\bibitem{wang2023data}
Wang, J.T., Jia, R.: {Data Banzhaf: A Robust Data Valuation Framework for Machine Learning}. In: AISTATS. PMLR, vol.~206, pp. 6388--6421 (2023)

\end{thebibliography}
\end{document}